\pdfoutput=1
\documentclass[10pt,twocolumn]{article}

\usepackage{wacv}
\usepackage{times}
\usepackage{graphicx}
\usepackage{amsmath}
\usepackage{amssymb}
\usepackage{makecell}
\usepackage{enumitem}
\usepackage{multirow}
\usepackage[numbers]{natbib}
\usepackage{xurl}

\usepackage[pagebackref=true,breaklinks=true,colorlinks,bookmarks=false]{hyperref}

\def\wacvPaperID{1130} 

\wacvfinalcopy 

\ifwacvfinal
\else
\fi

\begin{document}

\title{Face Verification with Challenging Imposters and Diversified Demographics}

\author{Adrian Popescu\textsuperscript{1}, Liviu-Daniel \c{S}tefan\textsuperscript{2}, Jérôme Deshayes-Chossart\textsuperscript{1}, Bogdan Ionescu\textsuperscript{2}\\
\textsuperscript{1}Université Paris-Saclay, CEA, List, F-91120, Palaiseau, France\\
\textsuperscript{2}University Politehnica of Bucharest, Romania\\
{\tt\small \{adrian.popescu,jerome.deshayes-chossart\}@cea.fr,\{liviu\textunderscore daniel.stefan,bogdan.ionescu\}@upb.ro}
}

\maketitle
\thispagestyle{empty}

\begin{abstract}
Face verification aims to distinguish between genuine and imposter pairs of faces, which include the same or different identities, respectively. The performance reported in recent years gives the impression that the task is practically solved. Here, we revisit the problem and argue that existing evaluation datasets were built using two oversimplifying design choices. First, the usual identity selection to form imposter pairs is not challenging enough because, in practice, verification is needed to detect challenging imposters. Second, the underlying demographics of existing datasets are often insufficient to account for the wide diversity of facial characteristics of people from across the world. To mitigate these limitations, we introduce the $FaVCI2D$ dataset. Imposter pairs are challenging because they include visually similar faces selected from a large pool of demographically diversified identities. The dataset also includes metadata related to gender, country and age to facilitate fine-grained analysis of results. $FaVCI2D$ is generated from freely distributable resources. Experiments with state-of-the-art deep models that provide nearly 100\% performance on existing datasets show a significant performance drop for $FaVCI2D$, confirming our starting hypothesis. Equally important, we analyze legal and ethical challenges which appeared in recent years and hindered the development of face analysis research.
We introduce a series of design choices which address these challenges and make the dataset constitution and usage more sustainable and fairer. 
$FaVCI2D$ is available at~\url{https://github.com/AIMultimediaLab/FaVCI2D-Face-Verification-with-Challenging-Imposters-and-Diversified-Demographics}.
\end{abstract}


\vspace{-6mm}
\section{Introduction}
Face verification (FV)
is deployed in applications such as biometrics \cite{chen2018mobilefacenets, labati2016biometric}, social media information structuring~\cite{hazelwood2018applied} or classical media archive organization~\cite{best2014unconstrained}. 
Performance has strongly progressed in recent years due to the introduction of deep learning techniques~\cite{8614364,10.1145/954339.954342}. 
Results reported for public datasets collected from heterogeneous sources, such as LFW~\cite{LFWTechUpdate}, IJB-C~\cite{maze2018iarpa}, MegaFace~\cite{kemelmacher2016megaface} or TrillionPairs~\cite{trillion}, are getting very close to 100\% accuracy. 

We find that such performance is misleading due to two design choices made when building existing face verification datasets. Following~\cite{phillips2012good} and~\cite{wang2019racial}, we hypothesize that the usual random selection of identities for imposter pairs makes the verification process too easy. 
Challenging imposter pairs in $FaVCI2D$ are created by using a deep face recognition model to get visually similar imposters. 
Then, a manual verification is done to ensure that the two identities are actually different. 
Challenging genuine pairs
are equally interesting to test the limits of verification systems. 
They are created by asking annotators to select images of the same identity which are visually different.

A second problem is the scale of the pool of imposters and the fairness of the face verification process. 
Imposter pool size is not central for the random selection of imposters if the pool is not too low. 
This parameter becomes important for challenging imposters since the dataset should contain faces with similar demographics for each target identity. 
Diversified and balanced demographic distribution of imposters is equally central to ensure a fair verification process~\cite{grother2019face}. 
If imposter demographics is imbalanced, a bias that is inverse compared to face recognition appears. 
A larger imposter pool from a given demographic split makes the verification process more challenging. 
The more imposters there are, the larger the chances of finding a similar one for a target identity. 
Note that scale was addressed in MegaFace~\cite{kemelmacher2016megaface} or DiF~\cite{merler2019diversity}, while fairness was only partially addressed in smaller datasets such as FairFace~\cite{karkkainen2019fairface} or BFW~\cite{robinson2020face}.
$FaVCI2D$ addresses these technical challenges. 
Imposter pairs are challenging because they include visually similar faces selected from a large pool of diversified identities. 
The dataset also includes demographic metadata to facilitate fine-grained analysis of results.

Legal and ethical aspects are central in face verification because the task deals with sensitive information related to subjects' identity. 
Ignoring or minimizing such aspects probably contributed to the withdrawal of datasets~\cite{van2020ethical}, such as MS-CELEB1M~\cite{guo2016ms}, MegaFace~\cite{kemelmacher2016megaface} and DiF~\cite{merler2019diversity}.
Legal criteria were considered during the constitution of $FaVCI2D$, notably in terms of enforcing copyright, personal data protection, and image rights. 
Reuse requirements are tackled by exploiting only resources which are released under suitable licenses.
Data protection regulations are different across the world. 
A recent comparison between the US, the EU and China's approaches to data protection~\cite{pernot2020china} concludes that the highest level of protection is offered by EU's General Data Protection Regulation (GDPR)~\cite{gdpr}. 
This is particularly the case for sensitive data, such as faces.
Consequently, the latter regulation should be taken into account when building face-related datasets.
Ethical aspects are also key in order for face verification and recognition to be developed in a socially acceptable way. 
Some of the concerns that have been voiced address: 
\vspace{-2mm}
\begin{itemize}[noitemsep]
\item Unlawful and/or unethical dataset collection ~\cite{megapixels} with problems such as: disclosure of the names of the persons, inclusion of copyrighted images and insufficient handling of consent, especially for children.
\item Bias against demographic categories, such as gender, age or origin~\cite{buolamwini2018gender,karkkainen2019fairface,robinson2020face,wang2019racial}. This is notably an effect of a strongly imbalanced collection of the face recognition datasets used to create deep models. 
\item Banning face recognition for law enforcement~\cite{abolishLetter} and surveillance~\cite{franceBan,canadaBan}. These initiatives are part of a larger debate regarding risks related to AI technologies. The raised objections should be carefully considered both from technical and legal perspectives to improve the acceptability of face-related technologies. 
\end{itemize}
\vspace{-2mm}
Our work is informed by these concerns and effort is devoted to addressing them. 
During dataset design, we put focus on: (1) compliance with legal requirements, (2) data minimization by storing only information necessary to fulfill the task, and (3) reducing demographic imbalance to ensure a fair analysis of demographic segments. 
Note that $FaVCI2D$ is built for face verification and, given the reduced number of faces per identity and anonymization of its identities, would be of no use to build recognition models directly. 
$FaVCI2D$ could be used during the construction of future recognition datasets to better calibrate them in terms of demographic representativity.
It is also noteworthy that, while the focus is put on security-related applications, face verification is useful in a range of other contexts. 
For instance, is increasingly used to organize large multimedia collections and thus improve access to their content~\cite{hazelwood2018applied,best2014unconstrained}.

\begin{table*}
\begin{center}
\resizebox{0.99\textwidth}{!}{

\begin{tabular}{ |c|c|c|c|c|c|c|c|c|c|c|c| }
\hline

Dataset & \makecell{$LFW$\\~\cite{LFWTechUpdate}} & \makecell{$YTF$\\~\cite{wolf2011face}} & \makecell{$IJB-C$\\~\cite{maze2018iarpa}}& \makecell{$MegaFace$\\~\cite{kemelmacher2016megaface}}& \makecell{$Trillion$\\$Pairs$~\cite{trillion}}& \makecell{$DiF$\\~\cite{merler2019diversity}}& \makecell{$FairFace$\\~\cite{karkkainen2019fairface}}& \makecell{$RFW$\\~\cite{wang2019racial}}& 
\makecell{$IJB-C{ext}$\\~\cite{chalap}} &
\makecell{$FaVCI2D$\\(proposed)}\\
\Xhline{3\arrayrulewidth}

\makecell{$\mathcal{T}_{1.1}$ genuine} & random & random & random & random &random & random & random & challenging & random & challenging \\
\hline
\makecell{$\mathcal{T}_{1.2}$ imposter} & random & random & random & random &random & random & random & challenging & random & challenging \\
\hline
\makecell{$\mathcal{T}_{2.1}$ unique IDs} & 5,749  & 1,595 & 3,531 & \makecell{530\\+distractors} & \makecell{5,749\\+distractors} & NK & NK & 3,000 & 6,139 & \makecell{12,468\\+distractors} \\ 
\hline
\makecell{$\mathcal{T}_{2.2}$ total images} & 13,233 & 621,000 & 31,334 & \makecell{1,000,000} & \makecell{1,580,000} & 1,000,000 & 108,501 & 1,000 & 152,917 & 64,879 \\ 
\hline
\makecell{$\mathcal{T}_{3.1}$ gender (F/M)} & 22.5/77.5 & NK & 37.3/62.7 & 41.1/58.9 & 22.5/77.5 & 48/52 & \makecell{50/50} & 35/65 & 38.5/61.5 & 44/56 \\
\hline
\makecell{$\mathcal{T}_{3.2}$ origin} & \makecell{race\\imbal.} & \makecell{race\\imbal.} & \makecell{race\\imbal.} & \makecell{race\\imbal.} & \makecell{race\\imbal.} & \makecell{skin color\\imbal.} & \makecell{race\\bal.} & \makecell{race\\bal.} & \makecell{skin color\\imbal.} & \makecell{country\\imbal.} \\
\hline
\makecell{$\mathcal{T}_{3.3}$ age} & NK & NK & estimated & NK & estimated & estimated & estimated & estimated & actual  & actual \\ 
\hline
\makecell{$\mathcal{T}_{4}$ ID type} & notable & notable & notable & any & notable & any & any & notable & notable & notable \\ 
\hline
\makecell{$\mathcal{L}_{1}$ reusable} & no & no & yes & yes & no & yes & yes & no & yes & yes \\ 
\hline
\makecell{$\mathcal{L}_{2}$ privacy} & no & no & no & no & no & no & no & no & no & yes \\
\hline
\makecell{$\mathcal{L}_{3}$ image rights} & yes & no & yes & no & no & no & no & yes & yes & yes \\
\hline
\makecell{$\mathcal{L}_{4}$ sustainability} & yes & yes & yes & no & yes & no & yes & yes & yes & yes \\
\hline
\end{tabular}
}
\end{center}
\label{tab:fv_analysis}
\vspace{-0.5em}
\caption{Overview of face verification datasets according to desirable characteristics. Genuine ($\mathcal{T}_{1.1}$) and imposter pairs ($\mathcal{T}_{1.2}$) are either random or challenging. Counts are provided for unique IDs ($\mathcal{T}_{2.1}$) and for images ($\mathcal{T}_{2.2}$), with distractors mentioned where used. Gender ($\mathcal{T}_{3.1}$) is the proportion of female/male images. Origin ($\mathcal{T}_{3.2}$) is given using race, skin color or country. Origin-related imbalance is mentioned. Age ($\mathcal{T}_{3.3}$) is estimated manually or automatically or actual when the photo was taken. ID type ($\mathcal{T}_{4}$) is either notable or any. Legal and ethical characteristics ($\mathcal{L}_{1}$ - $\mathcal{L}_{4}$) are described using yes/no . NK stands for ``not known" when information was unavailable.}
\vspace{-3mm}
\end{table*}

\section{Analysis of Face Verification Datasets}
\label{ref:existing}
Verification performance reported on widely-used datasets, such as LFW~\cite{LFWTechUpdate}, IJB-C~\cite{maze2018iarpa}, MegaFace~\cite{kemelmacher2016megaface}, is close to 100\%.
Such performance suggests that the task is solved or nearly so. 
However, the authors of~\cite{wang2019racial} show that if challenging imposter pairs are introduced in face verification, performance drops significantly.
This design choice makes face verification more realistic, and we build upon it in our work.
Given the sensitiveness of the task, much attention was given recently to different biases. 
The influence of gender, age, and ethnic origin was discussed, among others, in~\cite{buolamwini2018gender,karkkainen2019fairface,robinson2020face,wang2019racial} and should be carefully considered. 
With~\cite{terhorst2021comprehensive}, we note that non-demographic factors also introduce biases in face verification and deserve investigation.

\subsection{Face Verification Design Criteria}
We analyze datasets taking into account technical, legal and ethical aspects. 
We propose the following technical characteristics for a sound design of verification datasets:
\vspace{-6mm}
\begin{itemize}[noitemsep]
    \item $\mathcal{T}_1$: Development should be guided by real-life usage of this technology. The inclusion of hard genuine and imposter pairs ($\mathcal{T}_{1.1}$ and $\mathcal{T}_{1.2}$, respectively) is important to challenge the evaluated deep representations.
    \item $\mathcal{T}_2$: The number of identities ($\mathcal{T}_{2.1}$) and of images ($\mathcal{T}_{2.2}$) should be large enough to approximate large-scale face verification systems. 
    \item $\mathcal{T}_3$: A balanced spread of identities in terms of demographic factors such as gender ($\mathcal{T}_{3.1}$), geographic origin ($\mathcal{T}_{3.2}$) and age ($\mathcal{T}_{3.3}$) should be achieved. 
    \item $\mathcal{T}_4$: Datasets should include  identities which are representative for the general population, notable\footnote{\url{https://en.wikipedia.org/wiki/Wikipedia:Notability\_(people)}} or not. 
\end{itemize}
\vspace{-2.5mm}
We devise the following legal or ethical characteristics:
\vspace{-2.5mm}
\begin{itemize}[noitemsep]
    \item $\mathcal{L}_1$: FV datasets should be built on top of resources whose licenses allow reuse and modification. The use of other raw data was shown to be problematic in the long run and led to the withdrawal of some datasets. 
    \item $\mathcal{L}_2$: Compliance with data protection regulations is needed for a lawful distribution and usage of the dataset.
    A comparative analysis of data protection laws~\cite{pernot2020china} concludes that the EU's GDPR offers the highest level of protection for sensitive data. \textit{Art. 9} forbids the processing of such data, with exceptions for research in \textit{art 9.2(j)} if proportionality with the aim pursued is established as described in \textit{art. 89}. 
    \item $\mathcal{L}_3$: Compliance with other privacy laws, particularly with the image rights applicable in countries such as Canada, Belgium, France, or Spain. This right basically forbids the distribution of images that include recognizable faces. An exception is made for public figures when they appear in a professional capacity. 
    \item $\mathcal{L}_4$: Sustainability of the dataset is crucial for its future use. Some of the datasets were withdrawn because they generated strong debates about their adherence to legal and ethical standards~\cite{megapixels}.
\end{itemize}

The simultaneous optimization of all technical and legal criteria is difficult, if not impossible.
For instance, it is desirable to have a large number of identities in the dataset ($\mathcal{T}_2$).
However, the availability of identity images varies for factors such as gender and geographic origin~\cite{wagner2016women}.
This factor limits fairness in terms of geographic spread ($\mathcal{T}_3$ is targeted).
Equally important, the total number of available images ($\mathcal{T}_2$) is larger than that of reusable images ($\mathcal{L}_1)$ but the use of copyrighted images is risky.
We make the best effort to fulfill as many criteria as possible.

In Table~\ref{tab:fv_analysis}, we analyze nine existing FV datasets and $FaVCI2D$. 
We include technical and legal/ethical characteristics, which we deem important for sustainable and uncontroversial use of datasets. 
Controversies have negative impact on the public perception of face verification and recognition and ultimately hinder their development. 

\subsection{Analysis of Technical Criteria}
The first two characteristics compared in Table~\ref{tab:fv_analysis} refer to the way genuine and imposter pairs are created ($\mathcal{T}_{1.1}$ and $\mathcal{T}_{1.2}$, respectively).
As we mentioned, these criteria are important to ensure a realistic evaluation of face verification.
Challenging pairs are useful because verification in difficult conditions indicates how resilient the process is.
Given the very high level of performance reported on existing datasets such as LFW~\cite{LFWTechUpdate}, IJB-C~\cite{maze2018iarpa} or $IJB-C_{ext}$~\cite{chalap}, we estimate that the use of challenging genuine pairs is preferable to better compare the tested features.
The use of challenging imposter pairs creates a deception-oriented scenario in which each imposter pair attempts to fool the verification system.
We compare challenging and random imposters to point out differences between them.
Note that only RFW~\cite{wang2019racial} and $FaVCI2D$ include challenging genuine/imposter pairs which are created using visual similarities between IDs.
A related problem is that of the amount of curation applied to the included faces. 
Biometric verification works with a curated target (for instance, an ID photo) and a non-curated query image.
Other scenarios, such as verification in media archives or in social media, require verification in absence of curation since images in a pair come from uncontrolled sources.
Similarly to all recent datasets analyzed here, $FaVCI2D$ includes non-curated images.
The dataset is thus best fitted for usage for the second type of scenarios which gained a lot of traction.

The number of unique IDs ($\mathcal{T}_{2.1}$) used as probe is another important criterion insofar it enables a thorough evaluation. 
Existing datasets include thousands of identities to form genuine pairs. 
$FaVCI2D$ has the highest number of unique IDs (12,468) among the datasets for which this number is known. 
MegaFace~\cite{kemelmacher2016megaface} and Trillion Pairs~\cite{trillion} also include distractors to form a large number of diversified imposter pairs, as does $FaVCI2D$.
However, since selection is random for MegaFace and Trillion Pairs, the utility of using many distractors and a very large number of pairs is questionable. 
It would have been possible to further expand $FaVCI2D$ but at the price of increasing demographic imbalance, which is one of the main criticisms associated with existing datasets~\cite{buolamwini2018gender,karkkainen2019fairface,robinson2020face}.
The limitation comes from the number of notable persons belonging to underrepresented demographic segments, such as women from African countries.
This limitation is particularly strong when searching for pairs of representative and reusable images. 
Ongoing projects such as "Wiki loves women"\footnote{\url{https://en.wikipedia.org/wiki/Wikipedia:WikiProject\_Wiki\_Loves\_Women}} aim to reduce the demographic imbalance in Wikipedia.
We will later release revised versions of the dataset to reflect such changes.
The effect of the number of unique imposter IDs and of the number of unique IDs is evaluated in the experimental section.

The total number of images ($\mathcal{T}_{2.2}$) varies a lot across datasets. 
Megaface and Trillion Pairs include the highest number of images since they exploit a very large number of distractors. 
$YTF$~\cite{wolf2011face} was built from videos and it was easy to generate a large number of frames.
$FaVCI2D$ has fewer images because we choose to use only two images per ID to form genuine pairs and one image for imposter IDs.
This choice is made to enforce data minimization (see $\mathcal{L_2}$).

Gender distribution ($\mathcal{T}_{3.1}$) is a known problem in face verification and recognition~\cite{buolamwini2018gender}.
LFW, IJB-C, MegaFace, Trillion Pairs, RFW, and $IJB-C_{ext}$ are strongly imbalanced.
Efforts toward gender parity were made for DiF and FairFace.
In $FaVCI2D$, we wanted to diversify geographic spread and were able to achieve balance for Asia, America and Europe.
Unfortunately, reusable data was scarce for Africa.
We chose to match the number of IDs of African origins with those from other regions at the expense of strict gender parity for this region.

Origin ($\mathcal{T}_{3.2}$) is another highly sensitive criterion for which it is difficult to propose an objective and uncontroversial segmentation.
A majority of datasets use the notion of race to group people on this criterion.
However, the concept of race is, to say the least, controversial~\cite{lee2008ethics}.
Its use has also strongly contributed to the controversies which led to the withdrawal of face recognition datasets~\cite{van2020ethical}.
Following~\cite{chalap} and~\cite{merler2019diversity}, we decided to discard it in $FaVCI2D$.
Skin color~\cite{merler2019diversity} is more objective than race but we decided against its use since: (1) the same skin tone can characterize people of different origins or polyethnic combinations, (2) it can greatly vary due to the conditions in which a photo was taken and, more anecdotally, to tanning. 
Instead, we group people by their country of origin.
We acknowledge that this segmentation is equally imperfect because: (1) many countries have borders that do not correspond one-to-one to ethnic groups, (2) some people have multiple citizenship, and (3) a large part of people are polyethnic. 
However, this criterion is objective and less likely to lead to controversies than race or its approximation via skin color.

Age is the third important demographic criterion ($\mathcal{T}_{3.3}$) which should be tested in face verification. Some datasets have no age-related information, e.g., LFW, YTF, MegaFace, while others estimate it automatically, e.g., DiF, FairFace, RFW.
Following~\cite{chalap}, we decided to use the actual age of the persons when they were photographed. 
Since not all images have time-related metadata associated to them, we build subset of $FaVCI2D$ to estimate the effect of age.

The type of IDs included ($\mathcal{T}_{4}$) is another important criterion.  
While appealing, the use of faces of "common" people is legally and ethically difficult. 
Explicit consent would be needed from each person included for GDPR-compliance.
Such a constraint is highly impractical at a large scale but it can be waived for notable persons (see $\mathcal{L}_2$ and $\mathcal{L}_3$ below).  
Ethical challenges are equally strong because the inclusion of identities other than notable led to the withdrawal of $DiF$~\cite{merler2019diversity} and $MegaFace$~\cite{kemelmacher2016megaface}.
The type of included IDs should therefore be carefully considered during dataset design to ensure sustainable exploitation.
One interesting question that appears is whether the facial characteristics of notable persons are different from those of other people and thus affect the representativity of verification results.
Compared to existing datasets, the proposed demographic diversification reduces the probability for the retained sample faces to be different from the general population. 

\subsection{Analysis of Legal and Ethical Criteria}
The type of license associated with the images ($\mathcal{L}_{1}$) is a first important legal criterion. 
The inclusion of copyrighted images contributed to the withdrawal of MS-CELEB1M~\cite{van2020ethical} but datasets derived from it, including Trillion Pairs and RFW, are still distributed.
We note that a majority of analyzed datasets were created from reusable content.
Access to $FaVCI2D$ will be granted only after the signature of a contract which will specify the rights and obligations of the users, notably concerning exclusive use for scientific research purposes.

Data protection ($\mathcal{L}_{2}$) should be enforced when working with sensitive data such as faces.
Notably, data minimization stipulates that resources should only include information needed to carry out a technical task in a sound way.
Redundant information is included in most analyzed datasets. 
For instance, not all ID images from LFW are used to create pairs. 
$FaVCI2D$ complies with GDPR requirements on sensitive data because it is designed for research purposes (\textit{art. 9(j)}). It instantiates data minimization (\textit{art. 89.1}) by: (1) storing two images per identity which is the minimum quantity needed to perform verification, (2) removing the names of the persons from the dataset, (3) ensuring that all demographic segments include a sufficient number of identities. 
The following data-related rights are implemented to comply with \textit{art. 89.2}: right to access (\textit{art. 15}), right to rectification (\textit{art. 16}), right to restriction of processing (\textit{art. 18}) and right to object (\textit{art. 21}). A Web form through which any of the persons included in the dataset can require the expression of their rights will be made available. 
The proposed data protection measures establish proportionality between the proposed usage of data and the rights of the data subjects. 
To our knowledge, there is no publicly available legal analysis of GDPR compliance for face-related datasets.  
However, the proposed measures are in line with the recommendations made for the processing of genetic data~\cite{staunton2019gdpr}, which also fall in the sensitive category defined by GDPR in \textit{art. 9}. 

Image rights ($\mathcal{L}_{3}$) are respected if only notable people in public situations are included.
The repurposing of large datasets, such as YFCC100M~\cite{thomee2016yfcc100m}, led to the creation of MegaFace, DiF and FairFace.
While technically tempting, it is legally challenging in a range of countries.
For instance, the inclusion of childrens' faces in MegaFace, underlined in~\cite{megapixels}, is problematic. To be on the safe side, no children's faces were used in $FaVCI2D$.
The use of Wikipedia images minimizes the risk of including photos taken in private contexts, but a manual verification of the photos was still performed to exclude this risk.

Last but not least, sustainability ($\mathcal{L}_{4}$) is a core criterion for dataset usefulness. 
MegaFace and DiF were already withdrawn and Trillion Pairs, FairFace or RFW might be next due to public pressure.
The set of measures presented above should ensure long-term availability of $FaVCI2D$.

\section{Proposed Dataset}

\subsection{Identity selection and processing}
The first step of the data collection is to create a list of diversified identities with metadata associated to them.
Similarly to existing face recognition or verification datasets~\cite{cao2018vggface2,guo2016ms,LFWTechUpdate}, $FaVCI2D$ includes notable people. 
Differently from them, we aim to create a demographically diversified dataset by systematically exploiting metadata.

Identities are selected using a series of filters associated to demographic factors.
The first filter is related to age.
Only articles which are categorized under ``$YYYY$ births" in Wikipedia are kept, with $1920 \leq YYYY \leq 2000$.
The first bound is set to cover a wide variety of ages, while the second is used to avoid including images of children.

Geographical spread is the second filter used. 
Wikipedia entries are biased toward Western European, North American and other populous countries such as India and Japan~\cite{popescu2010spatiotemporal}.
This bias is reduced by imposing a maximum number of 10,000 entries per country for inclusion in $FaVCI2D$. 
Country names and associated demonyms for 183 countries are first searched in the article categories. 
If several of them occur, the one with the maximum count in the entire article is retained.
We checked country attributions for 500 identities and they were 99.4\% accurate.

Gender balance is targeted at the regional scale (Africa, America, Asia and Europe). 
This choice limits the total number of identities available in the dataset because Wikipedia exhibits a strong bias toward mens' biographies~\cite{wagner2016women}.
A classifier derived from~\cite{bamman2014unsupervised}, which counts the occurrences of the third-person singular feminine (she, her) and masculine (he, him, his) pronouns, is applied.
The authors of~\cite{bamman2014unsupervised} report 100\% accuracy and a verification of 500 identities from $FaVCI2D$ confirms their conclusion.

\subsection{Image collection}
The Bing Images search engine is used to download images for Wikipedia entries selected as described above. 
A first download includes up to 150 images per identity without license-related restriction.
This image set is used to create a visual representation of identities used only for pairs validation, as described below.
A second download collects up to 10 images with a reusable license per identity.
Photo metadata are exploited to estimate the age of the candidate faces. If EXIF data are available, the date when the photo was taken is exploited. 
Otherwise, year mentions are searched in URLs, photo titles and the HTML ALT descriptions.
Age is computed as the difference between the automatically obtained year of the photo and the birth year of the person.
This estimated age is manually verified during the pair selection process.

\subsection{Image preprocessing and reranking}
\label{subsec:rerank}
We aim to create a visual prototype for each identity in order to guide manual pair verification.
Face detection is applied to all downloaded images using MTCNN~\cite{xiang2017joint}.
Then, features are extracted for the detected faces using the $ir50$ model from~\cite{zhao2018towards} and L2-normalized.
Third, a mean ID feature is computed from the first 10 images returned by Bing in which only one face was detected.
This condition is necessary because it is initially impossible to know which one of the several faces in an image is relevant.
We assume that the top Bing results are on average more relevant than the following ones. 
Finally, all images of the identity are compared with the mean representation.
In parallel, a similarity matrix is computed among all the identities of the dataset using the mean representations made of top-10 ranked images.
This matrix is exploited to create $FaVCI2D$ variants used in the evaluation. 
Face features are also extracted for reusable images of the identity and are compared to the mean representations of all candidate identities.
Only faces that are closer to the current identity than to any other identity are retained for validation.

\subsection{Validation of image pairs}
This process is done in two steps.
An interface is created to select genuine pairs (see the interface in the supplementary material). 
A first annotator is instructed to select two of the reusable images which are relevant for the target identity.
Feedback is provided in the interface about the difference of age for the pairs which were already selected. 
Whenever several reusable images are available, the selection of candidates is guided by two related criteria: (1) the faces should be visually different and (2) the age difference should belong to one of the underrepresented bins is favored. 
Age difference is important insofar the faces change over time, but was not studied previously. 
A second interface is created for further verification of pairs.
Genuine pairs selected during the first step are checked by two more annotators.
They are kept only if both agree that the faces represent the same person.
The verification of challenging imposters is also done by three annotators to ensure that images of the same ID were not mistakenly kept in the pair.

\subsection{\textbf{\textit{FaVCI2D}} characteristics}
The proposed dataset includes identities from 153 countries.
30 countries of the initial 183 were excluded because they are heavily underrepresented.
The distribution of genuine identities for different countries is provided in the supplementary material. 
The total number of unique IDs is 52,411, with 12,468 of them being used in genuine pairs.
The total number of images is 64,879, with two images for IDs from genuine pairs and one for the imposter-only IDs.
The complete versions of $FaVCI2D$, created with random and challenging imposter selection, include a total of 24,936 pairs divided equally between the two types of pairs.

We target a balanced gender and geographic distribution. It was possible to obtain enough pairs for America, Asia and Europe but not for Africa. The dataset includes 3,708 genuine pairs, 50\% female - 50\% male, for each of the first three regions and 1,344 for Africa, 23.3\% female - 76.7\% male. 
The gender distribution of IDs in the entire dataset is 44\% female - 56\% male, which is the closest we can get with reusable resources to a perfect balance.

Age-related information was found for 6,535 out of a total of 12,468 genuine pairs. 
The distribution of ages at the moment when the photo was taken is: 17\% for 18-25 years old; 27\% for 26 - 35; 18\% for 36-45; 15\% for 46-55; 12\% for 56-65; 7\% for 66-75 and 4\% for 76 and over.
The distribution of age difference (in years) between two photos in genuine pairs is 18.5\% for the same age, 29.5\% for 1 and 2, 23\% between 3 and 5; 17\% between 6 and 10; and 12 for more than years. 
While relatively imbalanced the two age-related distributions, they include enough examples in each range to run an age-oriented analysis of verification results.

\begin{table}[!htb]
\begin{center}
\resizebox{\linewidth}{!}{
\begin{tabular}{|l|l|l|l|}
\hline
\textbf{Model} & \textbf{Training Data}           & \textbf{LFW} & \textbf{YTF} \\ \hline
\textit{insightface}    & MS-Celeb1M-ArcFace     & 99.87 (99.80+)       & 97.94        \\ \hline
\textit{ir152} & MS-Celeb1M & 99.76 (99.80) & 97.50 \\
\hline
\textit{seqface}        & MS-Celeb1M + Celeb-Seq & 99.80  (99.80)      & 98.00 (98.00)        \\ \hline
\textit{vgg}            & MS-Celeb1M + VGGFace2  & 99.40        & 96.78        \\ \hline
\textit{facenet}        & VGGFace2                & 99.55        & 95.12        \\ \hline
\end{tabular}
}
\end{center}
\vspace{-0.5em}
\caption{Accuracy (\%) of feature extractors on LFW and YTF.}
\label{tab:eval_models}
\end{table}

\begin{table*}[!htb]
\begin{center}
\resizebox{\textwidth}{!}
{
\begin{tabular}{|c|c|c|c|c|c|c|c|c|c|c|c|c|c|c|c|c|c|c|c|c|}
\hline
\textbf{Model} & \multicolumn{5}{c|}{\makecell{Similar = 1\\Imposter pool size}} & \multicolumn{5}{c|}{\makecell{Similar = 10\\Imposter pool size}} & \multicolumn{5}{c|}{\makecell{Similar = 100\\Imposter pool size}} & \multicolumn{5}{c|}{\makecell{Similar = random\\Imposter pool size}}\\
\hline

    & 1000 & 5000 & 10000 & 30000& 52410 & 1000 & 5000 & 10000 & 30000& 52410 & 1000 & 5000 & 10000 & 30000& 52410 & 1000 & 5000 & 10000 & 30000& 52410\\
\hline

\textit{insightface} &  \makecell{97.37\\ $\pm$ 0.03}  &  \makecell{96.76\\ $\pm$ 0.04}  &  \makecell{96.56\\ $\pm$ 0.03}  &  \makecell{96.07\\ $\pm$ 0.03}  &  \makecell{95.75\\ $\pm$ 0.0}  &  \makecell{98.04\\ $\pm$ 0.02}  &  \makecell{97.58\\ $\pm$ 0.05}  &  \makecell{97.34\\ $\pm$ 0.02}  &  \makecell{96.89\\ $\pm$ 0.05}  &  \makecell{96.64\\ $\pm$ 0.0}  &  \makecell{98.62\\ $\pm$ 0.02}  &  \makecell{98.22\\ $\pm$ 0.01}  &  \makecell{98.05\\ $\pm$ 0.03}  &  \makecell{97.77\\ $\pm$ 0.05}  &  \makecell{97.50\\ $\pm$ 0.0}  &  \makecell{98.81\\ $\pm$ 0.02}  &  \makecell{98.78\\ $\pm$ 0.01}  &  \makecell{98.82\\ $\pm$ 0.01}  &  \makecell{98.81\\ $\pm$ 0.03}  &  \makecell{98.82\\ $\pm$ 0.0} \\
\hline

\textit{ir152} &  \makecell{93.62\\ $\pm$ 0.03}  &  \makecell{92.10\\ $\pm$ 0.07}  &  \makecell{91.36\\ $\pm$ 0.05}  &  \makecell{90.08\\ $\pm$ 0.04}  &  \makecell{89.48\\ $\pm$ 0.0}  &  \makecell{95.52\\ $\pm$ 0.05}  &  \makecell{94.11\\ $\pm$ 0.08}  &  \makecell{93.43\\ $\pm$ 0.15}  &  \makecell{92.33\\ $\pm$ 0.05}  &  \makecell{91.84\\ $\pm$ 0.0}  &  \makecell{97.17\\ $\pm$ 0.02}  &  \makecell{96.13\\ $\pm$ 0.07}  &  \makecell{95.54\\ $\pm$ 0.02}  &  \makecell{94.64\\ $\pm$ 0.05}  &  \makecell{94.00\\ $\pm$ 0.0}  &  \makecell{97.65\\ $\pm$ 0.03}  &  \makecell{97.66\\ $\pm$ 0.03}  &  \makecell{97.65\\ $\pm$ 0.03}  &  \makecell{97.63\\ $\pm$ 0.04}  &  \makecell{97.64\\ $\pm$ 0.0} \\
\hline

\textit{seqface} &  \makecell{91.73\\ $\pm$ 0.15}  &  \makecell{89.46\\ $\pm$ 0.15}  &  \makecell{88.47\\ $\pm$ 0.16}  &  \makecell{86.65\\ $\pm$ 0.17}  &  \makecell{85.61\\ $\pm$ 0.0}  &  \makecell{94.58\\ $\pm$ 0.08}  &  \makecell{92.37\\ $\pm$ 0.11}  &  \makecell{91.31\\ $\pm$ 0.14}  &  \makecell{89.76\\ $\pm$ 0.16}  &  \makecell{88.99\\ $\pm$ 0.0}  &  \makecell{97.18\\ $\pm$ 0.06}  &  \makecell{95.55\\ $\pm$ 0.02}  &  \makecell{94.56\\ $\pm$ 0.10}  &  \makecell{93.04\\ $\pm$ 0.10}  &  \makecell{92.28\\ $\pm$ 0.0}  &  \makecell{98.04\\ $\pm$ 0.03}  &  \makecell{98.03\\ $\pm$ 0.04}  &  \makecell{98.00\\ $\pm$ 0.04}  &  \makecell{97.94\\ $\pm$ 0.02}  &  \makecell{98.06\\ $\pm$ 0.0} \\
\hline

\textit{vgg} &  \makecell{91.52\\ $\pm$ 0.15}  &  \makecell{89.01\\ $\pm$ 0.13}  &  \makecell{87.79\\ $\pm$ 0.08}  &  \makecell{86.00\\ $\pm$ 0.11}  &  \makecell{85.28\\ $\pm$ 0.0}  &  \makecell{94.44\\ $\pm$ 0.10}  &  \makecell{92.13\\ $\pm$ 0.07}  &  \makecell{90.89\\ $\pm$ 0.17}  &  \makecell{89.19\\ $\pm$ 0.10}  &  \makecell{88.29\\ $\pm$ 0.0}  &  \makecell{97.27\\ $\pm$ 0.09}  &  \makecell{95.33\\ $\pm$ 0.08}  &  \makecell{94.44\\ $\pm$ 0.07}  &  \makecell{92.85\\ $\pm$ 0.07}  &  \makecell{91.92\\ $\pm$ 0.0}  &  \makecell{98.37\\ $\pm$ 0.04}  &  \makecell{98.35\\ $\pm$ 0.06}  &  \makecell{98.31\\ $\pm$ 0.05}  &  \makecell{98.32\\ $\pm$ 0.06}  &  \makecell{98.42\\ $\pm$ 0.0} \\
\hline

\textit{facenet} &  \makecell{89.74\\ $\pm$ 0.14}  &  \makecell{86.90\\ $\pm$ 0.12}  &  \makecell{85.44\\ $\pm$ 0.13}  &  \makecell{83.48\\ $\pm$ 0.08}  &  \makecell{82.61\\ $\pm$ 0.00}  &  \makecell{93.51\\ $\pm$ 0.09}  &  \makecell{90.39\\ $\pm$ 0.04}  &  \makecell{89.13\\ $\pm$ 0.11}  &  \makecell{87.07\\ $\pm$ 0.12}  &  \makecell{86.06\\ $\pm$ 0.0}  &  \makecell{97.13\\ $\pm$ 0.05}  &  \makecell{94.79\\ $\pm$ 0.08}  &  \makecell{93.52\\ $\pm$ 0.12}  &  \makecell{91.28\\ $\pm$ 0.07}  &  \makecell{90.11\\ $\pm$ 0.00}  &  \makecell{98.37\\ $\pm$ 0.07}  &  \makecell{98.37\\ $\pm$ 0.06}  &  \makecell{98.36\\ $\pm$ 0.06}  &  \makecell{98.36\\ $\pm$ 0.06}  &  \makecell{98.39\\ $\pm$ 0.0} \\
\hline
\end{tabular}
}
\end{center}
\vspace{-0.5em}
\caption{Verification accuracy with different models and configurations. ``Similar" gives the position of the imposter identity in the ranked list of similar identities w.r.t. the reference identity in each imposter pair. The smaller this number is, the more challenging verification will be. ``Random" is the classical configuration in which imposters are selected randomly. ``Imposter pool size" gives the number of unique identities among which an imposter can be selected. The higher this number is, the more challenging the verification will be. Each configuration was run five times and the average accuracy and associated standard deviation are reported.}
\label{tab:global_results}
\vspace{-3mm}
\end{table*}

\section{Experimental Validation}
\label{sec:eval}
Different variants of our dataset are tested depending on the objective of each experiment. The number of genuine and imposter pairs is balanced in all configurations. 
A thorough evaluation of five state-of-the-art face verification models is proposed. First, these feature extractors are evaluated on two existing datasets. Second, we compare the behavior of these models using challenging/random imposters and a variable size of the pool of imposter IDs. Third, we examine the relation between accuracy and gender. Fourth, we compare the results obtained for 20 countries from four major regions of the world. Finally, we present results obtained for different age ranges and age differences.

\subsection{Evaluation with existing datasets}

The following models were used in experiments: \textit{insightface}~\cite{deng2018arcface}, based on ResNet-150, trained on MS-Celeb1M dataset using ArcFace loss; \textit{ir152}~\cite{zhao2018towards}, based on ResNet-152, trained on MS-Celeb1M dataset using Focal loss;  \textit{seqface}~\cite{hu2018seqface}, based on ResNet-27 trained on MS-Celeb1M using the L2-SphereFace loss and fine-tuned on Celeb-Seq dataset;  \textit{vgg}~\cite{cao2018vggface2}, based on SE-ResNet-50 trained on MS-Celeb1M dataset and fine-tuned on VGGFace2 dataset using Softmax loss; \textit{facenet}~\cite{schroff2015facenet}, based on Inception ResNet, trained on VGGFace2 dataset using SoftMax loss.

In Table~\ref{tab:eval_models}, we present the results obtained with the five models on LFW~\cite{LFWTechUpdate} and YTF~\cite{wolf2011face}.
When available, the original model performance is reported in parenthesis. 
The results reproduced here are coherent with the original ones. 
This finding validates the fact that the feature extractors are configured correctly and their further comparison is fair.

\subsection{Influence of imposter selection}

In Table~\ref{tab:global_results}, we present the accuracy of feature extractors in different configurations of imposter pair selection. The similarity between IDs that form imposter pairs is varied between 1 (usage of the most similar imposter ID) to random, the usual verification scenario. The size of the pool of IDs from which imposters are selected is varied between 1,000 and 52,410, the total number of IDs in $FaVCI2D$. Globally, the best performance is obtained with \textit{insightface} and the lowest with \textit{facenet}. 
The use of challenging pairs reduces performance quite significantly.
\textit{insightface} is the only method whose accuracy is above 90\% in the most challenging settings, i.e., most similar imposter ID and largest pool of imposters. 
The use of challenging pairs allows a better separability compared to a random selection of imposters.
When an entire pool of imposters is used, performance with random selection only varies from 97.64\% (\textit{ir152}) to 98.82\% (\textit{insightface}). 
The corresponding variation for the most challenging setting (Similar = 1) is from 82.61\% (\textit{facenet}) to 95.75\% (\textit{insightface}). 

The imposter pool size has virtually no influence for random selection of imposters. 
This result indicates that very large distractor sets, such as proposed in Megaface~\cite{kemelmacher2016megaface} or Trillion Pairs~\cite{trillion}, are useless in the random configuration. 
Inversely, the imposter pool size influences performance when similar imposters are used. The fact that performance is reduced between 30,000 and 52,410 imposter IDs indicates that an even larger number of unique IDs would have been useful in $FaVCI2D$.
However, the performance drop tends to reduce when increasing the imposter pool size.
Consequently, the dataset provides a usable approximation of the very large-scale performance of the tested models. 

\begin{table}[!htb]
\begin{center}
\resizebox{0.49\textwidth}{!}
{
\begin{tabular}{|c|c|c|c|c|c|c|c|c|}
\hline
\textbf{Model} & \multicolumn{2}{c|}{Sim. = 1} & \multicolumn{2}{c|}{Sim. = 10} & \multicolumn{2}{c|}{Sim. = 100} & \multicolumn{2}{c|}{Sim. = random}\\
\hline

    & F & M & F & M & F & M & F & M\\
\hline

\textit{insightface} &  95.14  &  96.30  &  96.17  &  97.06  &  97.26  &  97.72  &  98.93  &  98.73 \\
\hline
\textit{ir}152 &  88.38  &  90.47  &  90.69  &  92.87  &  93.34  &  94.59  &  97.52  &  97.76 \\
\hline
\textit{seqface} &  84.69  &  86.43  &  88.05  &  89.83  &  91.51  &  92.96  &  98.12  &  98.00 \\
\hline
vgg &  84.34  &  86.12  &  86.98  &  89.46  &  90.94  &  92.79  &  98.31  &  98.52 \\
\hline
\textit{facenet} &  81.64  &  83.47  &  85.21  &  86.82  &  89.69  &  90.49  &  98.39  &  98.39 \\
\hline
\end{tabular}
}
\end{center}
\vspace{-0.5em}
\caption{Verification accuracy for gender with 52,410 imposters.}
\label{tab:gender_results}
\vspace{-3mm}
\end{table}

These findings confirm that face verification is still an open research problem, especially when challenging imposter pairs are presented to the system. It would be interesting to use an even larger imposter pool to measure how much of a further performance drop is observed in challenging configurations. However, a larger the number of imposters would come at the cost of significantly increasing the demographic imbalance of the dataset.

We run an ablation experiment to estimate the influence of unique IDs count in $FAVCI2D$.
We remove 50\% and 25\% of IDs and test feature extractors with $Similar = 1$ from 52410 imposters. 
Five random samplings are used and accuracy is averaged.
The obtained results, detailed in the supp. material, are well aligned with those of the full dataset from Table~\ref{tab:global_results}.
The maximum differences are observed for $vgg$ and reach 0.25\% (85.03\ for 50\% ablation vs 85.28\% for the full dataset) and 0.09\% (85.19\% for 25\% ablation vs 85.28\%).
This indicates that unique IDs count is sufficient for a global evaluation of performance.
However, an enrichment of the dataset remains interesting for the evaluation of different demographic segments.

\subsection{Influence of gender}

The results from Table~\ref{tab:gender_results} indicate that accuracy is globally lower for female face verification. The performance gap between genders is larger when more similar faces are used as imposters. We note that there is virtually no difference for random imposter selection. Female pairs are recognized marginally better for \textit{insightface} and \textit{seqface}, while the opposite is true for \textit{ir152} and \textit{vgg}. The gap is largest for \textit{vgg} and \textit{facenet}, reflecting gender distribution imbalance from the face recognition datasets used for training the feature extractors. VGGFace2~\cite{cao2018vggface2} has stronger gender imbalance compared to MS-CELEB1M~\cite{guo2016ms}.
These datasets include fewer female than male identities.
Also, male identities have a larger average number of images associated with them. 
It would be interesting to verify if gender bias subsists for a feature extractor trained with a gender-balanced dataset.
This question should be carefully studied by future face verification but is beyond the immediate scope here.

\begin{table*}[!htb]
\begin{center}
\resizebox{0.99\textwidth}{!}
{
\begin{tabular}{|c|c|c|c|c|c|c|c|c|c|c|c|c|c|c|c|c|c|c|c|c|c|c|c|c|c|c|c|c|c|c|c|c|c|c|c|c|c|c|c|c|c|}
\hline

\multirow{2}{*}{\textbf{Model}} & \multirow{2}{*}{\textbf{Sim.}} & \multicolumn{5}{c|}{Region = Africa} & \multicolumn{5}{c|}{Region = America} & \multicolumn{5}{c|}{Region = Asia} & \multicolumn{5}{c|}{Region = Europe}\\
\cline{3-22}

 &  & \rotatebox{75}{S. Africa} & \rotatebox{75}{Nigeria} & \rotatebox{75}{Egypt} & \rotatebox{75}{Ghana} & \rotatebox{75}{Tunisia} & \rotatebox{75}{US} & \rotatebox{75}{Canada} & \rotatebox{75}{Brazil} & \rotatebox{75}{Argentina} & \rotatebox{75}{Mexico} & \rotatebox{75}{India} & \rotatebox{75}{Japan} & \rotatebox{75}{S. Korea} & \rotatebox{75}{Philippines} & \rotatebox{75}{China} & \rotatebox{75}{UK} & \rotatebox{75}{France} & \rotatebox{75}{Germany} & \rotatebox{75}{Italy} & \rotatebox{75}{Ireland}\\

\hline

\multirow{2}{*}{insightface} & 1 & 95.65 & 95.27 & 93.35 & 93.23 & 92.40 & 95.97 & 96.21 & 95.78 & 97.01 & 95.51 & 96.26 & 93.04 & 93.75 & 95.67 & 95.13 & 97.17 & 97.51 & 97.03 & 96.80 & 97.92\\
\cline{2-22}
 & random & 97.90 & 98.56 & 96.84 & 98.79 & 94.11 & 98.81 & 98.55 & 98.84 & 99.25 & 100.0 & 98.84 & 98.69 & 99.00 & 98.63 & 99.17 & 99.40 & 99.66 & 99.31 & 99.96 & 99.48\\
\hline
\multirow{2}{*}{ir152} & 1 & 84.35 & 87.03 & 89.06 & 85.47 & 84.64 & 90.38 & 89.15 & 88.98 & 86.56 & 83.33 & 91.53 & 84.21 & 86.29 & 89.19 & 88.86 & 91.81 & 91.89 & 90.71 & 91.97 & 92.88\\
\cline{2-22}
 & random & 97.72 & 98.14 & 95.36 & 98.46 & 93.70 & 97.67 & 97.40 & 98.37 & 97.59 & 96.77 & 98.06 & 96.31 & 96.84 & 97.60 & 97.52 & 98.35 & 98.69 & 98.56 & 98.19 & 98.77\\
\hline
\multirow{2}{*}{seqface} & 1 & 78.05 & 86.28 & 80.29 & 84.73 & 79.73 & 86.47 & 86.02 & 87.22 & 82.17 & 81.89 & 88.67 & 81.40 & 82.42 & 83.77 & 83.34 & 87.38 & 87.83 & 87.97 & 86.64 & 88.30\\
\cline{2-22}
 & random & 98.18 & 98.93 & 95.47 & 100.0 & 95.83 & 97.97 & 97.71 & 97.26 & 97.01 & 97.84 & 98.44 & 97.63 & 98.07 & 97.94 & 99.44 & 98.45 & 98.52 & 98.17 & 98.94 & 97.96\\
\hline
\multirow{2}{*}{vgg} & 1 & 81.75 & 82.88 & 83.90 & 84.73 & 77.94 & 86.17 & 84.75 & 86.06 & 89.38 & 82.88 & 87.30 & 80.87 & 78.78 & 85.21 & 84.19 & 85.30 & 87.62 & 89.13 & 85.80 & 87.79\\
\cline{2-22}
 & random & 98.65 & 98.43 & 97.88 & 99.19 & 95.83 & 98.35 & 97.85 & 98.33 & 99.17 & 98.65 & 98.58 & 97.87 & 98.26 & 98.19 & 99.33 & 98.45 & 98.70 & 99.02 & 99.13 & 99.39\\
\hline
\multirow{2}{*}{facenet} & 1 & 77.54 & 80.88 & 84.82 & 80.32 & 79.41 & 83.04 & 82.16 & 83.14 & 82.75 & 79.65 & 84.66 & 77.45 & 76.77 & 82.51 & 81.64 & 83.33 & 85.36 & 87.22 & 82.24 & 86.14\\
\cline{2-22}
 & random & 98.79 & 99.11 & 96.50 & 98.79 & 97.38 & 98.20 & 97.86 & 97.54 & 98.75 & 99.91 & 98.74 & 97.78 & 98.37 & 98.12 & 99.20 & 98.68 & 98.85 & 99.29 & 97.81 & 98.90\\
\hline
\end{tabular}
}
\end{center}
\vspace{-0.5em}
\caption{Verification accuracy for country of origin. The five countries with most representatives in the dataset are presented for each major region included in the dataset. Results are reported for an imposter pool size of 52,410.}
\label{tab:country_results}
\vspace{-3mm}
\end{table*}

\subsection{Influence of origin}

We present results obtained for countries of origin in Table~\ref{tab:country_results}.
Mirroring the results from Table~\ref{tab:global_results}, the differences between methods are higher for challenging imposters.
A more meaningful comparison of feature extractors can be made with challenging imposters.
\textit{Insightface} is best for all countries with challenging imposters.
The average performance is best for Europe, followed by America, Asia and Africa. 
These results reflect the structure of the underlying face recognition datasets which are biased toward Europe and North America. 
A stronger under-representation of some Asian countries seems to occur in VGGFace2~\cite{cao2018vggface2} compared to MS-CELEB1M~\cite{guo2016ms}, since results for Asia are lower for \textit{vgg} and \textit{facenet}, the two VGGFace2-based models. Performance for American countries often sits between that for Europe and those for Asian and African countries. 
This is interesting insofar American countries include an important mix of populations from other continents.

Within each region, there can be important differences between countries from the same region, even when their inhabitants would be grouped in the same ``race" category in other verification datasets, such as RFW~\cite{wang2019racial} or FairFace~\cite{karkkainen2019fairface}. 
This is, for instance, the case for Nigeria and Ghana in Africa or Mexico and Argentina in America and Japan and China in Asia. The use of country also provides interesting insights into which countries are most under-represented in face recognition datasets used to create the feature extractors. 
For instance, performance is low for Tunisia, Japan and South Korea for all extractors tested. 

Globally, the analysis presented in Table~\ref{tab:country_results} comforts our choice to use the country as a proxy for origin rather than race or skin color which were used previously. It also provides more support to the relevance of using challenging imposters instead of random ones in face verification.

\begin{table}[!htb]
\begin{center}
\resizebox{0.99\linewidth}{!}
{
\begin{tabular}{|c|c|c|c|c|c|c|c|c|}
\hline
\textbf{Model} & \textbf{Sim.} & 18-25 & 26-35 & 36-45 & 46-55 & 56-65 & 66-75 & 76-100\\

\hline

\multirow{2}{*}{\textit{insightface}} & 1 & 95.47 & 96.49 & 96.74 & 97.19 & 97.77 & 97.69 & 96.42\\
\cline{2-9}
 & random & 98.82 & 99.19 & 99.16 & 98.96 & 99.25 & 98.98 & 99.28\\
\hline
\multirow{2}{*}{\textit{ir152}} & 1 & 86.84 & 90.30 & 92.31 & 92.99 & 93.77 & 94.00 & 93.57\\
\cline{2-9}
 & random & 97.14 & 97.73 & 97.99 & 98.60 & 98.98 & 98.58 & 98.66\\
\hline
\multirow{2}{*}{\textit{seqface}} & 1 & 84.34 & 86.42 & 88.85 & 89.66 & 91.37 & 89.18 & 87.92\\
\cline{2-9}
 & random & 97.74 & 98.20 & 97.95 & 98.75 & 99.03 & 98.51 & 98.73\\
\hline
\multirow{2}{*}{\textit{vgg}} & 1 & 85.97 & 87.27 & 88.43 & 88.63 & 88.33 & 86.43 & 83.43\\
\cline{2-9}
 & random & 98.66 & 98.54 & 98.69 & 98.94 & 99.09 & 98.58 & 99.03\\
\hline
\multirow{2}{*}{\textit{facenet}} & 1 & 82.51 & 84.16 & 85.44 & 86.63 & 87.25 & 82.91 & 80.25\\
\cline{2-9}
 & random & 98.70 & 98.48 & 98.34 & 98.88 & 99.06 & 98.09 & 98.63\\
\hline
\end{tabular}
}
\end{center}
\vspace{-0.5em}
\caption{Verification accuracy for age ranges for 52,410 imposters.}
\label{tab:age_results}
\vspace{-3mm}
\end{table}

\begin{table}[!htb]
\begin{center}
\resizebox{0.85\linewidth}{!}
{
\begin{tabular}{|c|c|c|c|c|c|c|}
\hline
\textbf{Model} & \textbf{Sim.} & 0 & 1-2 & 3-5 & 6-10 & 10+\\

\hline

\multirow{2}{*}{\textit{insightface}} & 1 & 97.24 & 97.10 & 96.59 & 96.40 & 95.40\\
\cline{2-7}
 & random & 98.96 & 99.29 & 99.05 & 99.04 & 98.85\\
\hline
\multirow{2}{*}{\textit{ir152}} & 1 & 91.76 & 91.23 & 90.91 & 91.40 & 90.82\\
\cline{2-7}
 & random & 98.25 & 98.21 & 97.71 & 98.16 & 97.80\\
\hline
\multirow{2}{*}{\textit{seqface}} & 1 & 89.42 & 88.74 & 86.91 & 87.59 & 84.77\\
\cline{2-7}
 & random & 98.41 & 98.53 & 98.21 & 98.31 & 97.77\\
\hline
\multirow{2}{*}{\textit{vgg}} & 1 & 90.77 & 88.34 & 87.17 & 85.99 & 80.97\\
\cline{2-7}
 & random & 99.13 & 99.06 & 98.60 & 98.56 & 97.86\\
\hline
\multirow{2}{*}{\textit{facenet}} & 1 & 89.56 & 86.16 & 83.76 & 82.77 & 77.40\\
\cline{2-7}
 & random & 99.11 & 98.98 & 98.46 & 98.37 & 97.41\\
\hline
\end{tabular}
}
\end{center}
\vspace{-0.5em}
\caption{Accuracy for age difference ranges with 52,410 imposters.}
\label{tab:agediff_results}
\vspace{-3mm}
\end{table}

\subsection{Influence of age}
Two criteria are used here. 
Table~\ref{tab:age_results} illustrates the influence of the mean age of faces from genuine pairs.
Again, the comparison of results for challenging imposters is more meaningful since differences between age ranges are higher.
\textit{insightface} and \textit{ir152} provide rather stable results across age ranges.
Larger differences are observed for the other methods. Low performance is obtained for IDs at the two ends of the age spectrum (18-25 and 76-100) which are likely to be under-represented in the face recognition models used. 
It would be useful to focus on including persons from the extreme age ranges in the underlying face recognition datasets in order to reduce age-related bias.

Table~\ref{tab:agediff_results} illustrates the influence of age difference in genuine pairs. 
Results for challenging imposters show an inverse correlation between increasing age difference and performance. 
Similar to age, the most stable results are obtained for \textit{insightface} and \textit{ir152}. 
Larger drops with increasing age difference are observed for \textit{vgg} and \textit{facenet}, while \textit{seqface} sits in the middle. 
Differences between models are at least in part due to the underlying datasets, with MS-CELEB1M providing stabler discriminatory power across age difference ranges compared to VGGFace2. 

\section{Conclusions}
We first provided a detailed analysis of face verification datasets which highlights their merits and limitations.
We gave attention to legal and ethical aspects which are often discussed only marginally.
Compliance with such aspects should contribute to better public acceptance of face verification and avoid controversies that led to the withdrawal of datasets such as MS-CELEB1M, MegaFace and DiF.

This analysis led to the introduction of $FaVCI2D$ whose objective is to mitigate, to the extent possible, the limitations of existing datasets while preserving much of their qualities.
Focus is put on ensuring wide demographic coverage and on including challenging genuine and imposter pairs.
Demographic diversity and balance are obtained for most, but not all of the countries.
This situation is due to the fact that raw input data are strongly biased toward some regions of the world.
However, the demographic spread, balance and level of detail in $FaVCI2D$ is better than that of existing face verification datasets.

Finally, we proposed a fine-grained performance analysis with five deep face recognition models.
The evaluation shows that model comparison is more meaningful when using challenging imposter pairs. 
It also provides interesting insights related to steps needed in order to build a fair face verification process.
A wide majority of observed biases are actually due to demographically imbalanced training data used to create face recognition models.
Beyond its direct use in verification, the proposed dataset could be used during the constitution of future recognition datasets in order to better calibrate them in terms of demographic coverage.

\vspace{5mm}

\textbf{Acknowledgment}
This work was supported by the European Commission under European Horizon 2020 Programme, grant number 951911 - AI4Media. It was made possible by the use of the FactoryIA supercomputer, financially supported by the Ile-de-France Regional Council.

{\small
\bibliographystyle{ieee_fullname}
\bibliography{ms}
}

\end{document}


\title{Supplementary Material for "Face Verification with Challenging Imposters and Diversified Demographics"}

\author{Adrian Popescu\textsuperscript{1}, Liviu-Daniel \c{S}tefan\textsuperscript{2}, Jérôme Deshayes-Chossart\textsuperscript{1}, Bogdan Ionescu\textsuperscript{2}\\
\textsuperscript{1}Université Paris-Saclay, CEA, List, F-91120, Palaiseau, France\\
\textsuperscript{2}University Politehnica of Bucharest, Romania\\
{\tt\small \{adrian.popescu,jerome.deshayes-chossart\}@cea.fr,\{liviu\textunderscore daniel.stefan,bogdan.ionescu\}@upb.ro}
}

\maketitle
\thispagestyle{empty}

\section{Introduction}

In this supplementary material, we provide:
\begin{itemize}
    \item{An illustration of random and challenging imposters.}
    \item{Details about the annotation task and the interface used to validate genuine pairs in $FaVCI2D$.}
    \item{The distribution of genuine identities across countries.}
    \item{The detailed results obtained after the ablation of a percentage of genuine identities.}
\end{itemize}

\section{Illustration of imposters}

\begin{figure}[!htb]
\begin{center}
\includegraphics[width=0.99\linewidth,trim={0cm 0cm 0cm 0cm}]{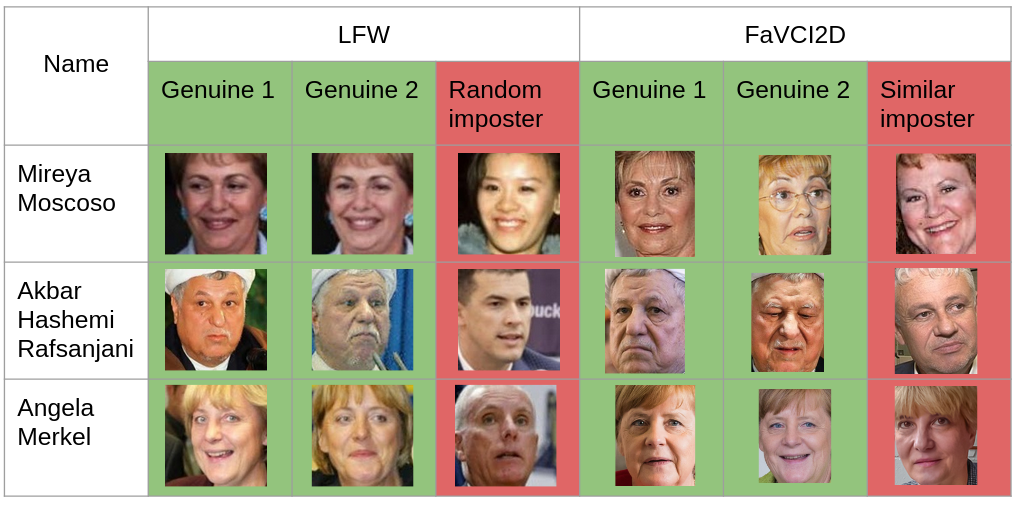}
\caption{Illustration of genuine and imposter pairs from LFW and $FaVCI2D$. LFW pairs are created randomly. Images in $FaVCI2D$ genuine pairs are selected as different from one another among available faces. $FaVCI2D$ imposters are samples of an identity which is visually similar to the genuine identity. }
\label{fig:overview}
\end{center}
\vspace{-3mm}
\end{figure}

Random and challenging imposters are presented in Figure~\ref{fig:overview} with examples from LFW and from our dataset proposed in this article, $FaVCI2D$ (Face Verification with Challenging Imposters and Diversified Demographics), respectively. 
The provided examples illustrate the fact that random imposters make the verification task too easy because the two faces from the random pairs are visually different from one another.
The selection of similar imposters provides a more realistic and challenging testbed for face verification.
Note also that the annotators are asked to choose challenging imposters whenever there are more than two faces available for an identity.

\section{Face annotation task}
\begin{figure*}[!htb]
\begin{center}
\includegraphics[width=0.99\textwidth,trim={0cm 0cm 0cm 0cm}]{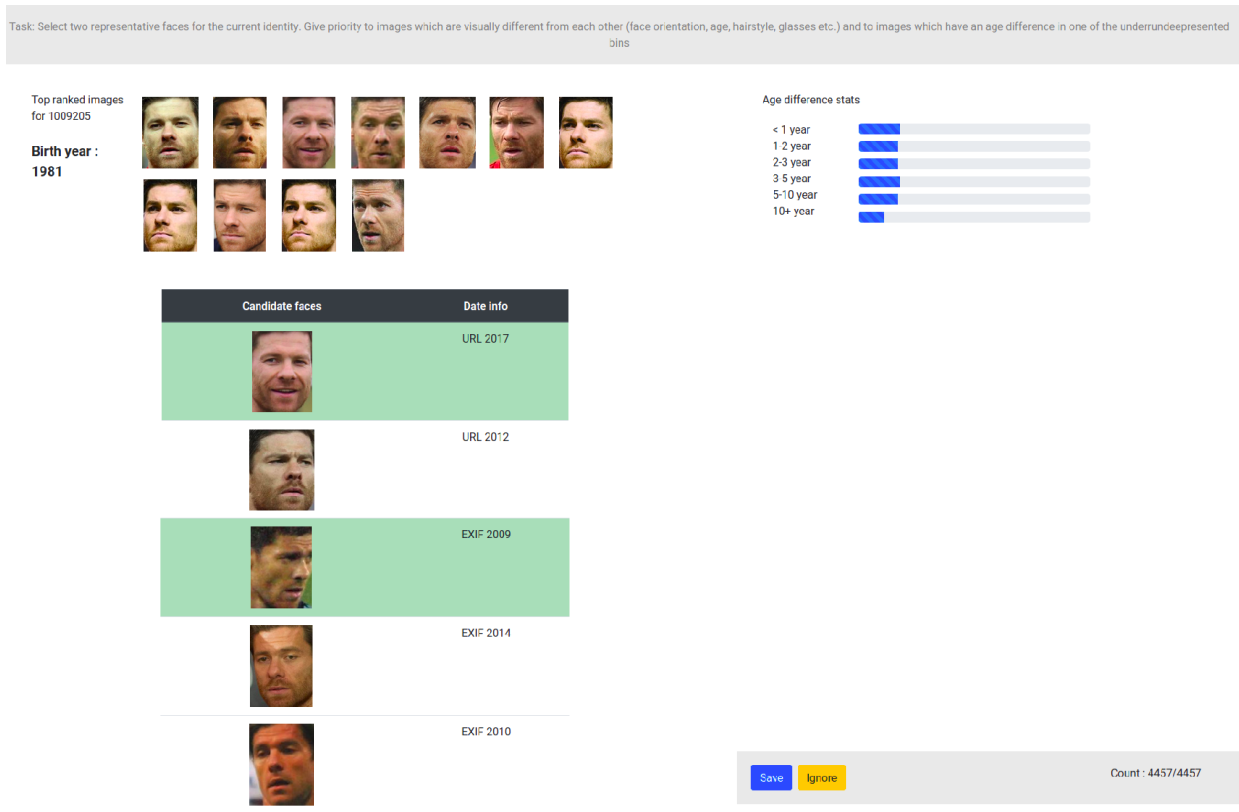}
\caption{Illustration of the interface used to collect visually challenging pairs for $FaVCI2D$.}
\label{fig:interface}
\end{center}
\end{figure*}

The annotation of genuine pairs was first performed by one participant.
Two versions of the interface were created depending on the availability of age related information for each identity (person).
We present the full instructions and interface,  which include age-related information, in Figure~\ref{fig:interface}.
A similar annotation process was implemented for genuine pairs which did not have age-related information associated to them. 
In this case, the age-related instructions and associated parts of the interface were naturally omitted. 

The participant who started the validation of genuine pairs received the following instructions: 
\textit{"Your participation is needed to create a dataset for face verification, a task whose objective is to automatically detect whether two faces represent the same identity or not. 
Each page of the interface contained the following elements:
\begin{itemize}
    \item Up to twelve reference faces which were automatically ranked as representative for the tested identity. These faces are presented in order to facilitate the evaluation of candidate faces. Note that, in some cases, a minority of reference faces might be irrelevant and they should be ignored. 
    \item The birth year of the person which is useful to evaluate if the age associated to candidate faces is realistic.
    \item At least two candidate faces with associated date taken information. The sources of the date taken information is indicated: EXIF, URL or ALT description. You should select two representative faces which, in addition of being relevant for the identity should: (1) be as different as possible from each other (face orientation, hairstyle, presence of glasses, beard etc.) and (2) have an age difference which is underrepresented in the bins to the right. Note that face selection is validated when its background turns to green. 
    \item Age-related statistics capture the number of identities in each age difference bin. The objective is, to the extent possible, to have a balanced representation of these difference. 
    \item The "Save" button allows you to validate a selected pair of faces and move to the next identity.
    \item The "Ignore" button allows you to skip the current identity if the candidate examples are are not representative of the identity. This can be due to: (1) uncertainty about their relevance for the current identity or (2) wrong assignment of the date taken information."
\end{itemize}
}

\begin{table}
    \centering
    \begin{tabular}{|c|c|c|c|}
     \hline
     \multicolumn{2}{|c|}{\textbf{Africa}} & \multicolumn{2}{|c|}{\textbf{America}} \\
     \hline
     South Africa  &  239 & United States  &  2100 \\
Nigeria  &  213 & Canada  &  396 \\
Egypt  &  97  & Brazil  &  322 \\
Ghana  &  83 & Argentina  &  267 \\
Tunisia  &  68 & Mexico  &  242 \\
     \hline
      \multicolumn{2}{|c|}{\textbf{Asia}} & \multicolumn{2}{|c|}{\textbf{Europe}} \\
     \hline
India  &  917 & United Kingdom  &  576 \\
Japan  &  673 & France  &  353 \\
South Korea  &  540 & Germany  &  341 \\
Philippines  &  216 & Italy  &  290 \\
China  &  209 & Ireland  &  249 \\
\hline
    \end{tabular}
    \caption{Distribution of the number of genuine pairs for  the most represented five countries in each region of the world included in $FaVCI2D$. The total number of genuine pairs is $12468$}
    \label{tab:countries}
\end{table}

The essential elements of these instructions were then reminded for each identity. 
The annotation interface deployed for the annotation of genuine pairs is illustrated in Figure~\ref{fig:interface}. 
In the example, the two selected faces differ in terms of face orientation and expression.
They are also distant in time since the first candidate photo was taken in 2017, while the second was taken in 2009. 
This age difference also contributes to the creation of a challenging genuine pair of images because faces tend to change over time.

Each genuine pair selected using the interface from Figure~\ref{fig:interface} is then verified by two more annotators using a simpler interface which displays only the preselected candidate faces and their age when this information is available. 
The two participants were instructed to validate pairs only if they were certain that both faces represented the same identity.
The pair is retained only if all three annotators agree that the two selected candidates are relevant.

The selection of most challenging imposters (Similar = 1 in Table 2 of the main paper) should be verified since the two identities can in some rare case be the same.
This situation can be an effect of the selection of: (1) fictional characters illustrated by images of the real-life person which is also included in the dataset or (2) from an erroneous selection of reference images for the identity.
To make sure that imposter pairs actually contained faces of different identities, we again displayed the pair of faces.
Three participants were asked to validate that the two candidate faces belonged to different identities. 
An imposter pair was retained only if the candidate faces were attributed to different identities by all three participants. 

\begin{table}[!htb]
\begin{center}
\resizebox{0.35\textwidth}{!}
{
\begin{tabular}{|c|c|c|c|}
\hline
\textbf{Model} & \multicolumn{3}{c|}{Ablation}\\
\hline

    & 50\% & 25\% & 0\% \\
\hline

\textit{insightface} & \makecell{95.69\\ $\pm$ 0.05} & \makecell{95.72\\ $\pm$ 0.04} & \makecell{95.75\\ $\pm$ 0.0}\\
\hline
\textit{ir152} & \makecell{89.43\\ $\pm$ 0.04} & \makecell{89.45\\ $\pm$ 0.04} & \makecell{89.48\\ $\pm$ 0.0}\\
\hline
\textit{seqface} & \makecell{85.67\\ $\pm$ 0.02} & \makecell{85.59\\ $\pm$ 0.02} & \makecell{85.61\\ $\pm$ 0.0} \\
\hline
\textit{vgg} & \makecell{85.03\\ $\pm$ 0.09} & \makecell{85.19\\ $\pm$ 0.07} & \makecell{85.28\\ $\pm$ 0.0} \\
\hline
\textit{facenet} & \makecell{82.54\\ $\pm$ 0.05} & \makecell{82.66\\ $\pm$ 0.02} & \makecell{82.61\\ $\pm$ 0.0} \\
\hline

\end{tabular}
}
\end{center}
\vspace{-0.5em}
\caption{Verification accuracy for the ablation of 50\% and 25\% of genuine IDs (pairs) from the dataset. Results with the full dataset (0\%) are also provided for reference. Results are reported with challenging imposters (Similar=1) selected among 52,410 IDs. The ablation was performed by randomly selecting five subsets of IDs and the reported results are averaged.}
\label{tab:ablation}
\vspace{-3mm}
\end{table}

\section{Distribution of genuine pairs across countries}
We discuss the distribution of genuine pairs a selection of top countries $FaVCI2D$.
The selected countries correspond those for which results were presented in Table 5 of the main paper. 
The distribution of identities per country is presented in Table~\ref{tab:countries}.
We note that while it was possible to ensure a balanced representation of each large region of the world, imbalance subsists for individual countries.
This is mainly an effect of the highly imbalanced character of the frequency of country-related identities in Wikipedia.
For instance, nearly a half of initial number of identities with associated photos were from the United States, a percentage which is reduced to less than 17\% of genuine pairs kept in $FaVCI2D$.
While surprising, the relatively low number of pairs from China is explained by the fact that this country has a highly skewed gender distribution in Wikipedia.
Nearly 90\% of Wikipedia identities with images describe Chinese men.
Inversely, it is somewhat surprising to note the relatively large number of Irish identities in the dataset.
This situation is probably explained by the fact that Ireland is an English-speaking country and has an active community of Wikipedia contributors.

\section{Results for ablated dataset}

The results from Table~\ref{tab:ablation} complete the ablation study mentioned in Subsection 4.2 of the main paper.
We ablate 50\% and 25\% of the total number of genuine identities from the datasets to verify that the number of unique IDs is sufficient.
The results confirm this hypothesis since the variation between the ablated datasets and its full version are small.
The maximum difference is obtained for \textit{vgg} when 50\% of IDs are removed and amounts to 0.25\%. 
The differences are under 0.1\% for all other models tested in Table~\ref{tab:ablation}.
Note also that, intuitively, the difference is smaller when the number of removed identities is 25\% compared to 50\%.